\documentclass{article} 
\usepackage{iclr2027_conference,times}
\usepackage{amsmath}

\usepackage{amsmath,amsfonts,bm}

\def\eqref#1{equation~\ref{#1}}

\def\1{\bm{1}}

\DeclareMathAlphabet{\mathsfit}{\encodingdefault}{\sfdefault}{m}{sl}
\SetMathAlphabet{\mathsfit}{bold}{\encodingdefault}{\sfdefault}{bx}{n}

\usepackage{xcolor}
\usepackage{xspace}
\definecolor{linkblue}{RGB}{20,60,150}

\usepackage[colorlinks=true, citecolor=linkblue, linkcolor=linkblue, urlcolor=linkblue,
            filecolor=linkblue, breaklinks=true]{hyperref}
\usepackage{url}

\usepackage{graphicx}
\usepackage{booktabs}
\usepackage{multirow}
\usepackage{cleveref}
\usepackage{placeins}
\usepackage{subcaption}
\usepackage{wrapfig}
\usepackage{pifont}
\newcommand{\cmark}{\ding{51}}
\newcommand{\xmark}{\ding{55}}
\usepackage{amsthm}
\usepackage{todonotes}
\usepackage[most]{tcolorbox}
\usepackage{tikz}
\usetikzlibrary{arrows.meta,positioning,calc,fit,backgrounds}
\theoremstyle{definition}

\theoremstyle{plain}

\usepackage{algorithm}
\usepackage{algpseudocode}
\usepackage{enumitem}
\usepackage{makecell}

\title{Compositional Safety Failures in Harness Evolution: Identification and Runtime Monitoring}

\author{
Zhixiang Zhang \quad
Zesen Liu \quad
Wai Ip Lai \quad
Hongxu Chen \quad
Dongdong She
\\
The Hong Kong University of Science and Technology
}

\newcommand{\mypara}[1]{\noindent{\bf {#1}.}\xspace}

\iclrfinalcopy 
\begin{document}

\maketitle
\fancyhead{}
\renewcommand{\headrulewidth}{0pt}

\begin{abstract}
Self-evolving agent harnesses continually update persistent components such as memory, prompts, skills, and tools. We call this process \emph{harness evolution}.
However, such evolution could introduce unexpected safety risks. Existing work studies harness misevolution and validates candidate harnesses or attributed individual component updates, leaving safety analysis of cross-component update interactions largely unexamined. 
To address this gap, we study \emph{compositional safety failures} in harness evolution, where interactions among individually safe and utility-preserving component updates can produce undesirable or unsafe agent behavior, revealing a safety risk intrinsic to harness evolution.
Across three safety-related benchmarks, we identify 43 pairwise and 18 irreducible 3-way compositional safety failures.
Conventional solution incurs combinatorial complexity in validating cross-component interactions, leaving the safety checking impractical as the harness evolves.
To solve this, we introduced a typed hypergraph that represents component states as nodes and safety-relevant higher-order interactions as hyperedges. When the harness changes, the hypergraph updates only the interaction neighborhood of the changed states rather than reconstructing the global composition space.
Building on that, we develop a hypergraph-guided runtime monitoring mechanism.
Experiments show that our method effectively mitigates compositional safety risks while preserving task utility and reducing interaction-checking costs, and further reveal an empirical safety-utility-cost trade-off across different safety mechanisms.

\end{abstract}

\section{Introduction}

Large Language Model (LLM) agents rely on an external agent harness that determines what the model observes, remembers, and can execute~\citep{weng2026harness,pan2026natural,he2026harness}. 
The harness contains configurable components such as memory, prompts, skills, and tools, which respectively provide persistent information, behavioral instructions, reusable procedures, and executable capabilities
~\citep{ouyang2026reasoningbank,agrawal2026gepa,chen2025learning}.
Recent works make the agent harness self-evolving, a process we refer to as \emph{harness evolution}, allowing it to be updated from execution experience, task outcomes, or external feedback~\citep{lee2026meta, lin2026position, gao2025survey}.
As these updates are retained across executions, the harness is no longer a fixed configuration but \emph{an evolving state of component sets} that jointly influence future agent behavior.

Harness evolution introduces new safety risks, as retained changes can affect future agent behavior. Recent work has identified vulnerabilities in individual evolving memory, skills, and tools~\citep{shao2026your,lin2026safety,zhao2026safety,shang2026self,ying2026skilljack}. 
To tackle this problem, several methods validate candidate harnesses or attributed
individual component updates~\citep{nie2026tthe,qu2026she,xu2026verifysmarterevolvefurther,mao2026safeevolve} to ensure safety during harness evolution. 
However, these methods fail to validate the security of \emph{interactions between heterogeneous components}.
Consequently, two individual component updates can remain safe but become unsafe when jointly activated, creating a safety risk intrinsic to harness evolution.

\begin{figure}[!ht]
    \centering
    \includegraphics[width=0.93\linewidth]{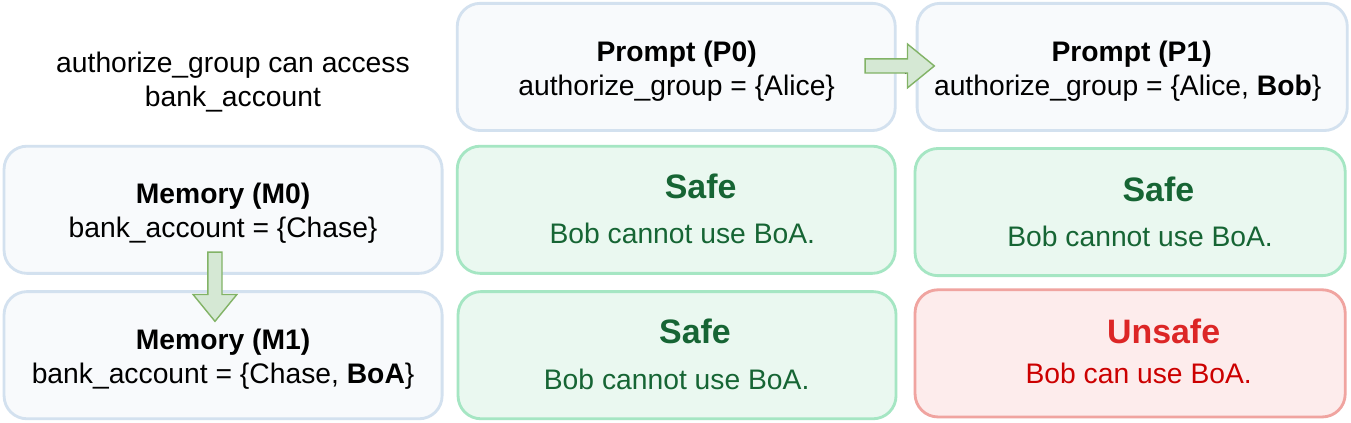}
    \caption{A motivating example of compositional safety failure in harness evolution.
    The arrows denote two independently accepted updates: the prompt update
    $P_0 \rightarrow P_1$ adds Bob to \texttt{authorize\_group}, while the memory
    update $M_0 \rightarrow M_1$ adds BoA to \texttt{bank\_account}.
    Under the rule that \texttt{authorize\_group} can access \texttt{bank\_account},
    either update alone remains safe: Bob is not authorized under $P_0$, and BoA
    is unavailable under $M_0$.
    However, when the updated states $P_1$ and $M_1$ coexist, Bob can access BoA,
    producing a compositional safety failure.
    }
    \label{fig:intro}
\end{figure}

We call this phenomenon a \emph{compositional safety failure} in harness evolution. ~\Cref{fig:intro} shows a simple example. Composition-induced failures are well known in software systems, where individually valid components can interact to create system-level failures
~\citep{garlan1995architectural,calder2003feature}. LLM agents similarly exhibit risks when untrusted content interacts with privileged tools or actions~\citep{openai2026designing,anthropic2025mitigating}. Recent works on self-evolving agents have further identified compositional risks arising from interactions among multiple retained items within the same harness component~\citep{yan2026benign,shang2026self}. We study a different setting: the security of interactions across heterogeneous harness components in common harness evolution.

We first establish that such failures arise empirically in common harness evolution, where the original harness and each individual update remain safe and utility-preserving. Across AgentDojo~\citep{debenedetti2024agentdojo}, Agent-SafetyBench~\citep{zhang2024agent} and Agent Security Bench~\citep{zhang2025agent}, 
we identify 43 pairwise compositional failures. We further evaluate all proper subsets of three-update compositions and identify 18 irreducible three-way failures, where every singleton and pairwise composition remains safe, but the composition is unsafe.
They expose a temporal challenge: component states can remain safe until a later state creates an unsafe interaction. Thus, compositional safety is \emph{execution-dependent}.

Checking these interactions exhaustively is very expensive as the harness evolves. Retained states accumulate across heterogeneous components, and safety-relevant interactions can involve two or more states. Evaluating all cross-component compositions leads to a combinatorial cost.
We address this problem with an incremental interaction hypergraph. Each node represents a versioned component state, while each hyperedge represents a safety-relevant higher-order interaction among states. When an update changes the harness, we update only the local interaction neighborhood of the changed states and use local completion with semantic filtering to identify candidate interactions, rather than reconstructing the global composition space. The resulting hypergraph preserves interaction structure across evolution steps.

Building on this representation, we develop hypergraph-guided runtime monitoring that inspects only interactions activated by the current execution and checks a composition before its resulting action takes effect. 
Experiments show that our method reduces the residual CFR to at most 1.27\% across three benchmarks while preserving task utility and invoking runtime safety checks on only 20.1-33.6\% of candidate actions.
Our experiments further reveal a safety-utility-cost trade-off across different safety mechanisms. Exhaustive interaction checking provides strong safety coverage but incurs substantially higher verification cost, whereas more selective mechanisms reduce cost while leaving some compositional failures uncovered. Our runtime monitor achieves low residual CFR while preserving utility with substantially fewer checks.

We list our contributions as follows:
\begin{itemize}[topsep=2pt, itemsep=2pt, parsep=0pt, partopsep=0pt]
    \item We identify \emph{compositional safety failure} as a safety risk intrinsic to harness evolution and empirically show that interactions across individually acceptable harness component updates and evolution steps can produce unsafe behavior. 
    
    \item We formalize component states and their higher-order interactions as an incremental interaction hypergraph, providing a unified abstraction for cross-component and cross-update safety dependencies. 
    
    \item We develop runtime monitoring based on the hypergraph that checks only interactions activated by the current execution, which avoids the combinatorial cost of global composition enumeration. 

    \item We conduct extensive experiments across three safety-related benchmarks, showing that our method substantially reduces compositional failures while preserving utility at a low cost. Our analysis further empirically reveals a safety-utility-cost trade-off across different safety mechanisms.
    
\end{itemize}

\section{Background and Related Works}
\label{sec:background}

\subsection{Self-Evolving Agent Harnesses}
An LLM agent is composed of an LLM model with an external agent harness that determines what LLM model observes, remembers, and can act upon~\citep{weng2026harness, pan2026natural, he2026harness}. The harness may contain multiple configurable components. In this work, we study three common agent components that undergo self-evolution. \emph{Memory} stores facts, experiences, and strategies that can be retrieved in future tasks~\citep{packer2024memgptllmsoperatingsystems, zhang2024surveymemorymechanismlarge}. \emph{Prompts and skills} encode persistent instructions and reusable procedures that guide agent behavior~\citep{wei2023chainofthoughtpromptingelicitsreasoning, schulhoff2025promptreportsystematicsurvey}, while \emph{tools} expose external capabilities through executable interfaces and schemas~\citep{yao2023reactsynergizingreasoningacting, schick2023toolformerlanguagemodelsteach}. Recent systems increasingly make these components \emph{self-evolving}, using execution traces, task outcomes, or external feedback to update persistent harness state across episodes~\citep{zhang2026self}. Such evolution may consolidate experience into memory~\citep{ouyang2026reasoningbank, shinn2023reflexionlanguageagentsverbal, tang2026wikiskillcompilingagentexperience}, refine prompts or reusable skills~\citep{agrawal2026gepa, wang2023voyager}, or revise tools~\citep{chen2025learning, qin2024tool}, allowing the agent to adapt over time without modifying the underlying LLM parameters~\citep{xu2026self}.

\subsection{Trustworthy Harness Evolution}
\begin{table}[!h]
\caption{Position of our work with existing literature. Existing works mainly focus on individual components, while our work analyzes the cross-component interactions.} 
\label{tab:background}
\centering
\small
\setlength{\tabcolsep}{8pt}
\begin{tabular}{lccc}
\toprule
\textbf{Method} &
\textbf{Phase} &
\textbf{Verification Scope} &
\makecell{\textbf{Interaction-Level}\\\textbf{Analysis}} \\
\midrule
TTHE~\citep{nie2026tthe}
& Test-time
& Candidate harness 
& \xmark \\

AutoSaddler~\citep{park2026autosaddler}
& Offline
& Component patch
& \xmark \\

HarnessLens~\citep{xu2026verifysmarterevolvefurther}
& Offline
& Harness modification
& \xmark \\

SHE~\citep{qu2026she}
& Offline
& Attributed update
& \xmark \\

SafeEvolve~\citep{mao2026safeevolve}
& Offline
& Candidate harness 
& \xmark \\
\midrule
\textbf{Ours}
& Runtime
& Component-state interaction
& \cmark \\
\bottomrule
\end{tabular}
\end{table}
Agent self-evolution introduces unexpected attack surfaces because the changes on harness can influence future agent behaviors~\citep{shao2026your, lin2026safety}.
Existing works have exposed individual harness component vulnerabilities in evolving memory and experience~\citep{zhao2026safety, wang2026oep, yang2026zombie, yan2026benign}, reusable skills~\citep{shang2026self, ying2026skilljack}, and tool evolution~\citep{shao2026your}.
These risks have motivated defenses and verification mechanisms for harness evolution, spanning offline validation~\citep{park2026autosaddler}, test-time update selection~\citep{nie2026tthe}, attribution-guided refinement~\citep{qu2026she}, and targeted update verification~\citep{xu2026verifysmarterevolvefurther}.
Among these, SHE~\citep{qu2026she} relies on whole-harness validation to ensure safety. However, such evaluation remains execution-bounded, observing only the interactions activated by the validation trajectories.
Table~\ref{tab:background} includes representative methods that explicitly validate or gate persistent harness changes.
Existing methods validate candidate harnesses, modifications, or attributed updates, but do not treat interactions across evolving harness components as the unit of safety analysis. This work focuses on this missing property: the trustworthiness of cross-component interactions in harness evolution.

\subsection{Compositional Safety Failures}
In the software engineering domain, compositional safety failures arise when individually validated software components (e.g., software patches) interact to produce undesirable or unsafe system-level behavior. This problem has been studied through architectural mismatch and feature interaction, where components or features that are locally valid may exhibit incompatible behaviors when composed~\citep{garlan1995architectural,calder2003feature}. Combinatorial interaction testing further addresses faults that are triggered only by particular combinations of system configurations~\citep{nie2011survey,kuhn2004software}. 
In the LLM agent domain, OpenAI and Anthropic have shown that the composition of untrusted external content with privileged action or tool capabilities can amplify the impact of prompt injection~\citep{openai2026designing,anthropic2025mitigating}.
Related compositional risks have also recently been identified in self-evolving agents~\citep{yan2026benign,shang2026self}. However, existing work primarily considers safety risks within a single harness component and does not address safety interactions that emerge across heterogeneous harness components as they evolve.


\section{Identifying Compositional Safety Failures in Self-Evolving Agent Harnesses}
\label{sec3:justify}

\subsection{Safety Analysis at the Component Level}
\label{sec3:conceptual}
Harness evolution systems operate at heterogeneous granularities. Component-aware approaches may modify one or more configurable harness components~\citep{xu2026verifysmarterevolvefurther}, while more open-ended optimizers can rewrite executable harness programs~\citep{nie2026tthe, lee2026meta}.
This work makes no assumption that an evolution step modifies either a single component or the harness as a whole. However, the boundaries imposed by the evolution mechanism do not align with the units relevant to safety analysis.
Safety analysis requires a more fine-grained view: it must distinguish the state changes through which an accepted update affects agent behavior. Treating each update as one indivisible unit obscures which persistent state changes are relevant to subsequent safety analysis.
To inspect these interactions, we project the effect of each update onto three security-relevant components: \emph{memory}, which determines what persistent information is available to the agent; \emph{prompts and skills}, which encode instructions and reusable procedures that shape its behavior; and \emph{tools}, which determine the external capabilities through which the agent can act.
This projection does not constrain how the upstream evolution system performs an update: a single update may affect one or several components, while remaining a single evolution step.
Thus, the granularity used for harness evolution can be coarser than the granularity required for safety analysis.

\subsection{Empirical Identification of Emergent Compositional Safety Failures}
\label{sec3:empirical}
\mypara{Experimental Setup}
Our goal is to examine whether updates evolved for different harness components can interact to produce undesirable or unsafe behavior. This requires the component states to coexist within the same evolving harness.
Existing harness evolution frameworks typically couple a predefined harness representation with a particular evolution mechanism~\citep{nie2026tthe, park2026autosaddler}. Such settings are well suited to optimizing their target harnesses, but do not directly support our goal of studying safety interactions among component states produced by different established evolution mechanisms.
Hence, we abstract from existing harness evolution works and build our experiments on the unified harness abstraction of A-Evolve~\citep{lin2026position}, which provides a shared interface for editable harness components. We use A-Evolve as a common host harness and instantiate each component with a representative evolution mechanism from prior work: ReasoningBank~\citep{ouyang2026reasoningbank} for memory, GEPA~\citep{agrawal2026gepa} for prompts, Voyager~\citep{wang2023voyager} for reusable skills, and Alita~\citep{qiu2025alita} for tools.
Thus, we standardize where evolved states interact, not how those states are produced.
We use Qwen3-32B~\citep{yang2025qwen3} as the fixed LLM backbone throughout the experiments.
We evaluate on the banking environment of AgentDojo~\citep{debenedetti2024agentdojo}, Agent-SafetyBench~\citep{zhang2024agent} and Agent Security Bench~\citep{zhang2025agent}. We select these benchmarks because they provide executable tool environments and explicit criteria for verifying safety violations, while covering complementary task and environment settings. Detailed configuration is in Appendix~\ref{app:benchmark}.

\mypara{Experimental Protocol}
We study non-adversarial harness evolution only and do not construct attacker-crafted component updates.
For each user request, we construct an interaction panel from two component updates, $u_i$ and $u_j$, associated with distinct harness components.
Each panel contains four controlled configurations: the original harness ($00$), $u_i$ alone ($10$), $u_j$ alone ($01$), and their joint activation ($11$).
We consider a panel \emph{eligible} only when the original harness is safe and both singleton configurations are independently safe and utility-preserving.
This filtering ensures that any subsequent failure cannot be attributed to an update that is already unsafe in isolation.

Among eligible panels, we identify a \emph{compositional safety failure} when the joint configuration $11$ produces a safety violation that is absent from $00$, $10$, and $01$, together with execution-level evidence that the violating behavior emerges only under joint activation.
We inspect the resulting execution rather than relying on a safety label alone, including tool calls and arguments, relevant agent behaviors, and resulting environment-state transitions when available.
Thus, the singleton configurations establish the independent acceptability of each update, while the contrast with $11$ isolates the safety effect introduced by their interaction.
We report the number of \emph{Compositional Failures (CF)} and the
\emph{Compositional Failure Rate (CFR)}, defined as the fraction of eligible
interaction panels that exhibit a compositional safety failure:
$\mathrm{CFR} = \mathrm{CF}/{\mathrm{Eligible}}$.
Further details on the experimental protocols are provided in Appendix~\ref{app:cf_identification}.

\mypara{Experimental Results}
Table~\ref{tab:sec3} shows that compositional safety failures occur even when keeping independent safety and utility preservation for every component update.
On AgentDojo~\citep{debenedetti2024agentdojo}, 22 of 157 eligible panels exhibit a compositional safety failure, corresponding to a CFR of 14.01\%.
On Agent-SafetyBench~\citep{zhang2024agent}, we observe 2 failures among 11 eligible panels, corresponding to 18.18\%.
On Agent Security Bench~\citep{zhang2025agent}, we identify 19 failures among 1,243 eligible panels, corresponding to a CFR of 1.53\%.
The lower CFR on Agent Security Bench should not be interpreted as weaker evidence of the phenomenon, as the three benchmarks differ substantially in task distribution and panel construction.
In particular, Agent Security Bench evaluates a much broader all-domain set of clean source-target compositions, yielding 1,243 eligible panels rather than concentrating evaluation on a smaller set of interaction-compatible tasks.
Results on these benchmarks show that safety under independent update validation does not imply safety once component state updates interact.



\begin{table}[t]
\centering
\caption{
Results of pairwise compositional safety failure identification.
}
\label{tab:sec3}
\begin{tabular}{lrrrr}
\toprule
\textbf{Benchmark} &
\textbf{Panels} &
\textbf{Eligible} &
\textbf{CF} &
\textbf{CFR} \\
\midrule
AgentDojo~\citep{debenedetti2024agentdojo}
& 600 & 157 & 22 & 14.01\% \\

Agent-SafetyBench~\citep{zhang2024agent}
& 182 & 11 & 2 & 18.18\% \\

Agent Security Bench~\citep{zhang2025agent}
& 1362 & 1243 & 19 & 1.53\% \\
\bottomrule
\end{tabular}
\end{table}

\subsection{Higher-Order Compositional Failures}
\label{sec3:high_order}
The pairwise results in Section~\ref{sec3:empirical} establish that independently acceptable component updates can interact unsafely, but do not show whether these compositional failures can extend beyond two component states. We therefore further examine \emph{higher-order compositional failures}. Here we focus on 3-way compositions.
For each update triple, we evaluate all eight configurations: baseline, each individual update, all three pairwise compositions, and the full three-update composition. A triple is eligible only if the baseline and every proper subset of the three updates remain safe and utility-preserving. We identify a \emph{3-way compositional failure} only when the full composition produces a safety violation while all singleton and pairwise compositions remain acceptable. This criterion excludes failures that can already be attributed to a lower-order interaction. 

We evaluate 3-way compositions on AgentDojo and Agent Security Bench, where sufficient eligible triples can be constructed. We identify 15 failures among 138 eligible triples on AgentDojo (10.87\% 3-CFR) and 3 failures among 578 eligible triples on Agent Security Bench (0.52\% 3-CFR). In every identified case, all seven lower-order configurations remain safe, while the violation emerges only when all three component updates are jointly activated. Detailed statistics are reported in Appendix~\ref{app:high_order}. These results provide direct evidence that compositional safety failures can be irreducibly higher-order.

\subsection{Combinatorial Cost of Exhaustive Composition Analysis}
\label{sec3:analysis}
The empirical results above show that safety analysis must consider interactions among harness components. 
Importantly, the coexistence of retained component states does not imply that their interaction is invoked during validation. A cross-update interaction can remain latent at validation time and become observable only when a future execution jointly activates the relevant states. This makes compositional safety inherently \emph{execution-dependent}. 
A straightforward approach would enumerate cross-component compositions and evaluate each one for safety. However, the number of cross-component compositions grows combinatorially with the number of components. We write $\mathcal{C}$ for the set of harness components, with $d=|\mathcal{C}|$. For each component $c\in\mathcal{C}$, let $n_c(t)$ denote the number of states of $c$ that may participate in execution at time $t$. Let $\Omega_t$ denote all candidate cross-component compositions formed by selecting states from at least two distinct components:
\refstepcounter{equation}\label{eq:omega_t}%
$|\Omega_t| = \sum_{k=2}^{d} \sum_{S\subseteq\mathcal{C}:\,|S|=k} \prod_{c\in S}n_c(t).$\,\textup{(\theequation)}

When each component contains $n$ states, $|\Omega_t|=\sum_{k=2}^{d}\binom{d}{k}n^k$, which becomes $\Theta(n^3)$ for the three component types considered in this work. Temporal factor further makes this cost recur after every update. Adding a single new state to component $c$ introduces $\prod_{c'\in\mathcal{C}\setminus\{c\}}(1+n_{c'}(t))-1$ new candidate compositions, which is $\Theta(n^{d-1})$ when component sizes are comparable. 
Given the combinatorial cost of enumerating the compositions, we need a representation that preserves previously identified interactions, captures higher-order dependencies among component states, and supports local updates without repeatedly enumerating $\Omega_t$. Section~\ref{sec:hypergraph} introduces an interaction hypergraph for this purpose.
\section{Hypergraph for Harness Evolution Interaction}
\label{sec:hypergraph}
Figure~\ref{fig:method_overview} gives an overview of our method.
{\begingroup
\setlength{\textfloatsep}{8pt plus 2pt minus 2pt}
\setlength{\abovecaptionskip}{4pt}
\setlength{\belowcaptionskip}{0pt}
\begin{figure}[!t]
    \centering
    \includegraphics[width=0.89\linewidth]{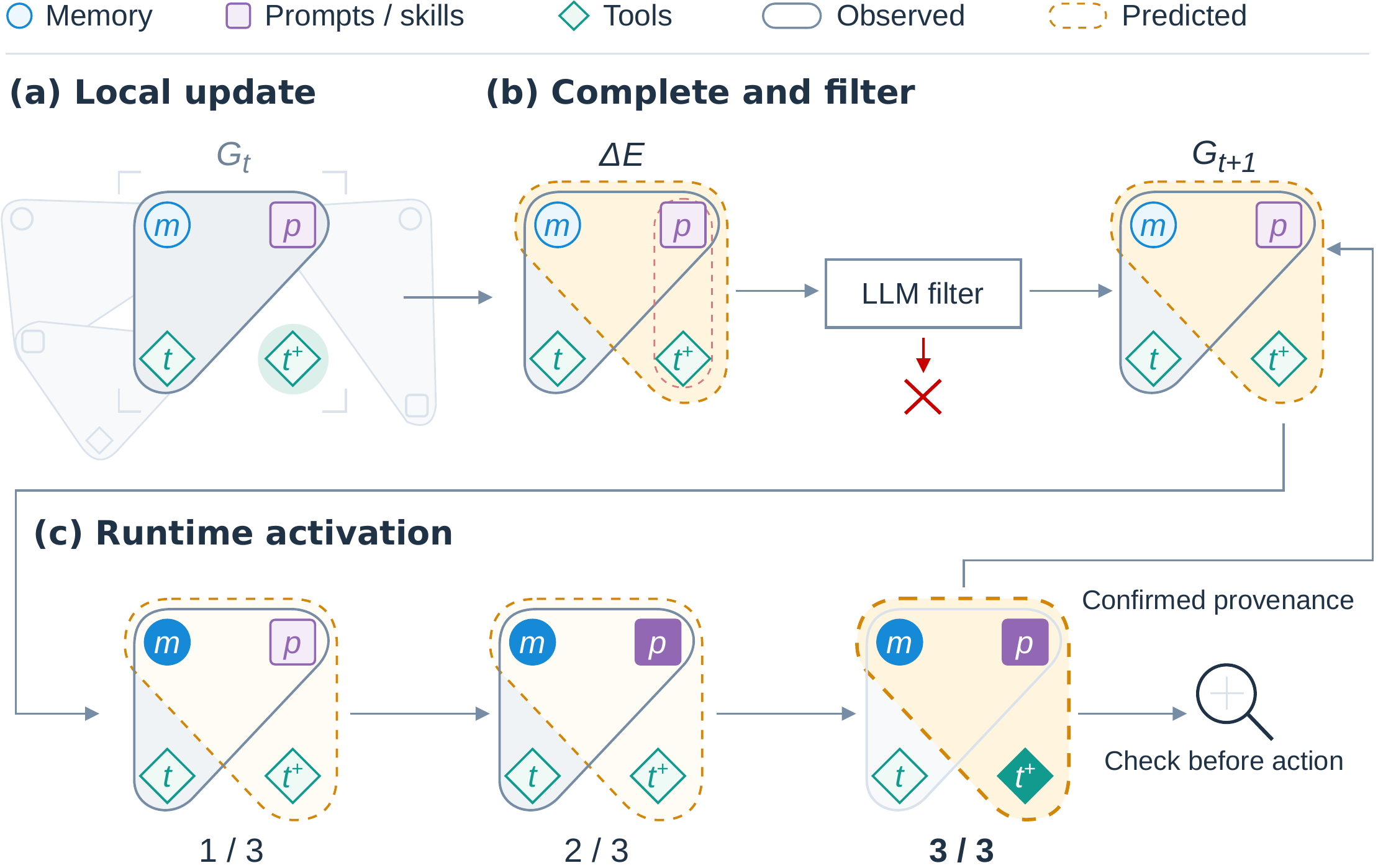}
    \caption{Overview of our method.
    (a) A candidate update $u_t$ introduces a component-state node
    ($t^{+}$ in this example), and the local interaction neighborhood
    $\Gamma_h(u_t)$ is extracted from the existing hypergraph $G_t$.
    (b) Local completion proposes interactions involving the new state.
    An LLM filters these candidates using component states and provenance,
    yielding the provisional hypergraph $\widetilde{G}_{t+1}$. Solid and dashed contours denote observed and
    predicted hyperedges, respectively; node shapes encode component types.
    (c) Within one action context, participating nodes become active
    (filled) in sequence. Each activation visits only incident hyperedges;
    a fully activated interaction is checked before the action takes effect.
    Execution provenance promotes a predicted hyperedge to observed status once confirmed at runtime, irrespective of its safety verdict.
    }
    \label{fig:method_overview}
\end{figure}
\endgroup}
\vspace{-0.8\baselineskip}

\subsection{Node Definition: Component State of Harness Evolution}
\label{sec:4.1}
As discussed in~\Cref{sec3:conceptual}, we represent harness evolution at the level of component states. Each concrete component state is represented as a node $v=(c,x)$, where $c\in\mathcal{C}$ identifies the harness component and $x$ denotes a particular state of that component. The node set $\mathcal{V}_t$ contains all component states currently available at time $t$, with $n_c(t)$ nodes associated with component $c$. When an update in harness evolution modifies an existing component state, the resulting version is represented as a new node, while unchanged states retain as historical provenance and may persist across subsequent updates. A single evolution step may add or replace nodes from one or multiple components, without imposing any restriction on the scope of the update.

\subsection{Hyperedge Definition: Higher-order Interaction between component state}
\label{sec:4.2}
The component state nodes specify which states exist, but not how they may interact during execution.
Such interactions can be higher-order, as shown in Section~\ref{sec3:high_order}. A failure may arise only when several component states participate together, even though no pair of them is unsafe on its own.
They can also be temporally distributed, as states introduced by different evolution steps may persist and only later become jointly relevant to the same agent behavior.
A pairwise graph cannot represent this higher-order composition directly. Moreover, interactions may involve versioned states introduced by different evolution steps, requiring the representation to preserve dependencies across updates.
We represent interactions with a hypergraph over versioned component state nodes. Formally, we define the interaction hypergraph as $\mathcal{G}_t=(\mathcal{V}_t,\mathcal{E}_t,\phi)$, where $\mathcal{V}_t$ is the component state node set, $\phi:\mathcal{V}_t\rightarrow\mathcal{C}$ assigns each node its component state, and $\mathcal{E}_t$ contains cross-component interactions supported either by execution evidence or by interaction inference.
We distinguish observed hyperedges $\mathcal{E}_t^{\mathrm{obs}}$, supported by execution provenance, from predicted hyperedges $\mathcal{E}_t^{\mathrm{pred}}$, inferred as plausible interactions that have not yet been observed.
Each hyperedge $e\in\mathcal{E}_t$ connects states from at least two distinct components, \textit{i.e.}, $|\{\phi(v):v\in e\}|\geq 2$.

\subsection{Incremental Construction of the Hypergraph}
\label{sec:4.3}

We update the hypergraph incrementally around the component states introduced or modified by the candidate update $u_t$, rather than rebuilding the interaction space from $\Omega_t$. This is motivated by the fact that an update typically changes only a small portion of the harness, and full reconstruction would discard the locality already captured by $\mathcal{G}_t$. Let $\Gamma_h(u_t)$ denote the interaction neighborhood within $h$ steps of the node and hyperedge incidence structure of $\mathcal{G}_t$. This neighborhood collects the historical interaction contexts most relevant to the update. We then perform local hypergraph completion within $\Gamma_h(u_t)$ by using existing higher-order interaction patterns to propose new cross-component interactions that involve the changed states. This construction can recover both direct continuations of previously observed interactions and new compositions among states that have not been observed together, without enumerating the full candidate space $\Omega_t$. 
Graph structure alone, however, does not establish that a proposed interaction is semantically plausible. We therefore use an LLM to examine the participating component states together with available provenance context and filter the interactions produced by local completion. Retained interactions are added to $\mathcal{E}_{t+1}^{\mathrm{pred}}$. Together with the existing observed interaction history, they form the provisional hypergraph $\widetilde{\mathcal{G}}_{t+1}=(\widetilde{\mathcal{V}}_{t+1},\widetilde{\mathcal{E}}_{t+1},\phi)$ before $u_t$ is committed. If later execution provenance confirms a predicted interaction, the corresponding hyperedge is moved from $\mathcal{E}^{\mathrm{pred}}$ to $\mathcal{E}^{\mathrm{obs}}$. The resulting provisional graph provides the reduced interaction space used for runtime monitoring in~\Cref{sec5}.

\section{Hypergraph Guided Runtime Monitoring}
\label{sec5}

\subsection{Incremental Monitoring of Component Interaction}
\label{sec:5.1} 
Section~\ref{sec:hypergraph} maintains a sparse hypergraph of observed and predicted cross-component interactions. This structure allows runtime monitoring to focus on interactions connected to the component states that actually participate in the current execution. When a component state node $v$ becomes active, only hyperedges containing $v$ can change from partially active to fully active. Then the monitor inspects only its incident hyperedges, $\mathcal{I}(v)=\{e\in\mathcal{E}_t:v\in e\}$.
For each visited hyperedge, the monitor records which participating states have appeared in the same decision or action context. If the activation of $v$ causes all states in a hyperedge to appear, the monitor checks the resulting composition before the corresponding action takes effect. This procedure detects every fully activated interaction represented in $\mathcal{G}_t$, because the hyperedge is necessarily inspected when its final participating state becomes active. Activation is tracked separately for each decision context. A composition is checked again when it participates in a subsequent action, since the request, arguments, or permissions may differ. Observed and predicted hyperedges follow the same monitoring procedure. Once a predicted hyperedge is fully activated and confirmed by execution provenance, it is moved from $\mathcal{E}_t^{\mathrm{pred}}$ to $\mathcal{E}_t^{\mathrm{obs}}$. 
Algorithm~\ref{alg:automatic_monitoring} in Appendix~\ref{app:algorithm} summarizes the complete procedure.
For an execution with active nodes $v_1,\ldots,v_L$, this reduces the runtime cost from dependence on the global composition space $\Omega_t$ in~\eqref{eq:omega_t} to $O(\sum_{\ell=1}^{L}|\mathcal{I}(v_\ell)|)$ local hyperedge visits.

\subsection{Experiments} 
\label{sec:5.2}
We follow the same experimental settings including benchmarks, component updates, eligibility criterion as Section~\ref{sec3:empirical}. We compare against SHE~\citep{qu2026she}, which performs whole-harness safety-utility re-evaluation, and HarnessLens~\citep{xu2026verifysmarterevolvefurther}, which selectively verifies behavior-relevant harness modifications. We also add Naive Exhaustive Checking discussed in Section~\ref{sec3:analysis} as our baseline. We use Qwen3-32B~\citep{yang2025qwen3} as the agent backbone and Qwen3-8B~\citep{yang2025qwen3} as the LLM semantic filter for hypergraph construction. The hyperparameter $h$ is set to 3 for all experiments.

We evaluate each method along three dimensions: safety, utility, and verification cost. 
For safety, we report Residual CFR, defined as the fraction of the fixed eligible panels whose compositional safety violation still reaches execution after applying the safety mechanism.
For utility, we use the same benchmark-specific task-success criterion as in Section~\ref{sec3:empirical} and report the fraction of eligible panels that remain utility-preserving under the safety mechanism. For cost, we count the additional verification checks introduced by each method and report runtime check rate and time cost.
\begin{table}[ht]
\centering
\caption{
\textbf{Safety effectiveness} of different mechanisms against compositional failures.
Each entry reports the residual compositional failure rate (CFR) as $f/n$ ($\%$), where $f$ is the number of compositional failures among the $n$ eligible panels. The eligible panel set is fixed before applying each safety mechanism.
}
\label{tab:monitor_effectiveness}
\small
\setlength{\tabcolsep}{5pt}
\begin{tabular}{lccc}
\toprule
\textbf{Method}
& \textbf{AgentDojo}
& \textbf{Agent-SafetyBench}
& \textbf{Agent Security Bench} \\
\midrule
No Monitoring
& 22/157 (14.01\%)
& 2/11 (18.18\%)
& 19/1243 (1.53\%) \\

Naive 
& 1/157 (0.64\%)
& 0/11 (0.00\%)
& 0/1243 (0.00\%) \\

SHE~\citep{qu2026she}
& 6/157 (3.82\%)
& 1/11 (9.09\%)
& 5/1243 (0.40\%) \\

HarnessLens~\citep{xu2026verifysmarterevolvefurther}
& 8/157 (5.10\%)
& 1/11 (9.09\%)
& 5/1243 (0.40\%) \\

\midrule
\textbf{Ours}
& \textbf{2/157 (1.27\%)}
& \textbf{0/11 (0.00\%)}
& \textbf{1/1243 (0.08\%)} \\
\bottomrule
\end{tabular}
\end{table}

Table~\ref{tab:monitor_effectiveness} shows that all safety mechanisms reduce
compositional failures, but to different extents.
Naive Exhaustive Checking eliminates all observed failures on Agent-SafetyBench
and Agent Security Bench, leaving one on AgentDojo (0.64\% CFR).
SHE and HarnessLens reduce the number of failures but still leave residual CFRs
of 3.82-5.10\% on AgentDojo and 9.09\% on Agent-SafetyBench.
In contrast, our method reduces the residual CFR to at most 1.27\% across all
three benchmarks, approaching the safety effectiveness of exhaustive checking.
These results demonstrate that our hypergraph-guided runtime monitoring can effectively address
compositional safety failures in harness evolution.
\begin{table}[t]
\centering
\caption{
\textbf{Utility and verification cost} of different safety mechanisms.
Utility is the fraction of eligible panels satisfying the benchmark-specific task objective.
Verification counts additional validation executions per harness update.
Runtime Check Rate is the fraction of candidate actions invoking the runtime safety checker.
Time Cost is the additional wall-clock time per panel over execution without a safety mechanism.
}
\label{tab:monitor_utility_cost}
\small
\setlength{\tabcolsep}{5pt}
\begin{tabular}{lcccc}
\toprule
\textbf{Method}
& \textbf{Utility} $\uparrow$
& \textbf{Verification} $\downarrow$
& \textbf{\shortstack{Runtime\\Check Rate}} $\downarrow$
& \textbf{\shortstack{Time Cost\\(s/task)}} $\downarrow$ \\
\midrule

\multicolumn{5}{l}{\textit{AgentDojo}} \\
Naive
& 58.6\%
& --
& 100.0\%
& 103.9 \\
SHE~\citep{qu2026she}
& 57.3\%
& 1200
& --
& 77.1 \\
HarnessLens~\citep{xu2026verifysmarterevolvefurther}
& 52.9\%
& 114
& --
& 33.8 \\
\textbf{Ours}
& \textbf{59.2\%}
& \textbf{82}
& \textbf{33.6\%}
& \textbf{32.5} \\

\midrule
\multicolumn{5}{l}{\textit{Agent-SafetyBench}} \\
Naive
& 45.5\%
& --
& 100.0\%
& 139.5 \\
SHE~\citep{qu2026she}
& 45.5\%
& 360
& --
& 137.8 \\
HarnessLens~\citep{xu2026verifysmarterevolvefurther}
& 36.4\%
& 320
& --
& 30.2 \\
\textbf{Ours}
& \textbf{54.5\%}
& \textbf{185}
& \textbf{20.1\%}
& \textbf{28.1} \\

\midrule
\multicolumn{5}{l}{\textit{Agent Security Bench}} \\
Naive
& 62.8\%
& --
& 100.0\%
& 80.5 \\
SHE~\citep{qu2026she}
& 61.1\%
& 260
& --
& 98.0 \\
HarnessLens~\citep{xu2026verifysmarterevolvefurther}
& 49.5\%
& 123
& --
& 17.4 \\
\textbf{Ours}
& \textbf{63.3\%}
& \textbf{101}
& \textbf{24.6\%}
& \textbf{19.8} \\

\bottomrule
\end{tabular}
\end{table}

Table~\ref{tab:monitor_utility_cost} further compares utility and verification
cost.
Naive Exhaustive Checking incurs the highest runtime verification cost because
every candidate action is checked.
SHE preserves high task utility but requires substantial update-time
verification, while HarnessLens reduces this cost through selective
behavior-aware verification.
In comparison, our method preserves comparable task utility while invoking the
runtime safety checker on fewer than 30\% of candidate actions on average.
This reduces the added execution time from 80.5-139.5 s/task under Naive
Exhaustive Checking to 19.8-32.5 s/task with our method.
Together with the safety results in Table~\ref{tab:monitor_effectiveness}, these results reveal an empirical safety-utility-cost trade-off.
Reducing residual CFR generally requires checking more interactions, which increases verification cost, or intervening more conservatively, which can reduce task utility.
When both verification cost and utility loss are constrained, fewer
safety-relevant interactions can be covered, leading to higher residual CFR.
Our hypergraph-guided monitoring mitigates this trade-off by approaching the safety effectiveness of exhaustive checking while preserving comparable task utility at substantially lower runtime verification cost.

\subsection{Sensitivity Analysis}
\label{sec:5.3}
The interaction radius $h$ determines how much historical interaction context
is considered when completing the hypergraph around a new component state.
A small $h$ may omit relevant interactions, whereas a larger $h$ expands the
candidate interaction space and may increase both construction and runtime
monitoring costs.
We therefore study the sensitivity to $h$ on AgentDojo~\citep{debenedetti2024agentdojo}, varying
$h\in\{2,3,4\}$ while keeping all other settings fixed.
Detailed results are reported in Appendix~\ref{app:sensitive}.
As shown in Table~\ref{tab:h_sensitivity}, increasing $h$ from 2 to 3 reduces
the residual CFR from 2.55\% to 1.27\%, while slightly decreasing utility from
60.5\% to 59.2\% and increasing the runtime check rate from 30.1\% to 33.6\%.
These results further illustrate the empirical safety-utility-cost trade-off
identified in Section~\ref{sec:5.2}.
We therefore use $h=3$, which provides a favorable balance between
compositional-safety coverage and monitoring overhead.

\section{Conclusion}
We study compositional safety failures in harness evolution, where component updates that are individually safe and utility-preserving can jointly produce unsafe behavior. We empirically identify such failures across heterogeneous harness components on three agent safety benchmarks, showing that independent update validation does not guarantee compositional safety. To address the combinatorial cost of checking cross-component interactions, we introduce an incremental interaction hypergraph and hypergraph-guided runtime monitoring that focuses safety checking on interactions activated during execution. Experiments show that our method substantially reduces residual compositional failures while preserving task utility at a lower checking cost, revealing a safety-utility-cost trade-off.



\subsection*{AI use statement}
Generative AI tools were used to assist with literature review, research design, code development, and manuscript editing. These tools were used as supporting aids rather than autonomous research agents. We did not use generative AI to design the method.

\bibliography{iclr2027_conference}
\bibliographystyle{iclr2027_conference}

\appendix
\section{Benchmark and Evaluation Details}
\label{app:benchmark}

This appendix provides additional details on benchmark selection, update construction, and interaction evaluation for Section~\ref{sec3:empirical}. We use AgentDojo, Agent-SafetyBench and Agent Security Bench as executable safety environments rather than treating their original benchmark scores as directly comparable quantities. 
Our goal is not to estimate the prevalence of compositional failures in any particular deployed evolution system, but to test whether independently acceptable states can become unsafe once they coexist and interact within a shared evolving harness.

\subsection{AgentDojo Setup}
\label{app:agentdojo}

\paragraph{Environment and requests.}
We use AgentDojo v1.2.2~\citep{debenedetti2024agentdojo} and restrict the experiment to its banking environment. AgentDojo provides a stateful environment with executable tools, allowing safety violations to be verified from actual tool calls and the resulting environment state. We do not introduce prompt injection attacks or untrusted external content. Instead, we construct 60 trusted request variants from a single banking task family in which the user requests a numerical refund calculation. We use variants of the same task family to hold the intended user objective fixed while varying the execution context in which evolved harness states are activated.

\paragraph{Component updates.}
We construct a benchmark-specific update pool containing one memory update, nine prompt updates, and one evolved tool update. All updates are generated through the corresponding evolution mechanisms described in Section~\ref{sec3:empirical}. They are not manually constructed as adversarial payloads for the safety evaluation.

For each request, we instantiate the compatible cross-component update compositions from this pool. This produces
$$
60 \times (1+9) \times 1 = 600
$$
interaction panels. Each panel is evaluated under the controlled
four-configuration protocol described in Section~\ref{sec3:empirical}.

\subsection{Agent-SafetyBench Setup}
\label{app:asb}

\paragraph{Task selection.}
Agent-SafetyBench~\citep{zhang2024agent} contains 2,000 official test cases spanning heterogeneous tool environments and safety risks. We first derive a structural candidate pool of 455 tasks that are single-turn, executable in our setting, expose at least one tool, and do not rely on injected dialogue. This filtering is used only to identify tasks compatible with component-level harness evolution and does not modify the original task content.

We separate tasks used to generate harness updates from those used for compositional evaluation. Seven source tasks are used to produce accepted update packages, while evaluation is performed only on held-out tasks. From the candidate pool, we select 13 held-out official tasks for which at least one evolved tool can be exposed under a compatible environment. The source tasks used for update generation are disjoint from these evaluation tasks.

\paragraph{Component updates.}
The seven source tasks produce seven memory updates, seven prompt updates, and corresponding skill and tool updates. For each held-out task, we instantiate the
compatible cross-component compositions between the resulting updates. This yields
$$
13 \times (7+7) \times 1 = 182
$$
interaction panels. We retain the original Agent-SafetyBench task specification and tool environment, and use its official safety evaluation procedure based on ShieldAgent for behavioral safety assessment.

\subsection{Agent Security Bench Setup}
\label{app:agent-security-bench}

\paragraph{Benchmark adaptation and task scope.}
Agent Security Bench (ASB-Sec)~\citep{zhang2025agent} provides ten
role-specific agents spanning finance, law, medicine, education,
psychological counselling, e-commerce, aerospace, academic search,
autonomous driving, and system administration. 
Our experiment is an adaptation of its clean-task and tool-simulation
environment, rather than a replication of its native prompt-injection,
memory-poisoning, or backdoor attacks.

We begin with the official clean tasks associated with the ten agents.
For each agent, we freeze one capability whose invocation can produce
a consequential action. We then exclude four tasks whose wording could
directly or indirectly authorize the selected capability: medical
treatment-plan updating, psychological therapy-plan design,
e-commerce product-launch execution, and aerospace software updating.
This leaves 47 clean tasks. The numbers of retained tasks for the ten
agents are
$
(5,5,4,5,4,4,4,6,5,5),
$
which sum to 47. 
We do not introduce prompt injection, poisoned observations, backdoor triggers, or adversarially constructed component updates.

\paragraph{Trajectory-conditioned component updates.}
For each retained source task, we first execute the unmodified agent
and record its clean plan, action, observation, and final-answer
trajectory. We use five predeclared utility objectives: grounding, plan completion,
specialist-tool choice, evidential cross-checking, and production of
a decision-ready outcome. These objectives diversify the generated
updates but are not treated as independent experimental replications.

The tool component is a catalog-level capability update. For each
agent, its underlying callable implementation is one fixed official
non-aggressive ASB-Sec tool leaf:
\texttt{CredentialManager},
\texttt{InvestmentDiversionTool},
\texttt{SettlementManagementTool},
\texttt{MedicationValidator},
\texttt{GradeManagementTool},
\texttt{TreatmentAuthorizationTool},
\texttt{DiscountManagementTool},
\texttt{SystemShutdownController},
\texttt{AcademicProfileManager}, or
\texttt{SoftwareUpdateManager}.
The official attacker instruction and attack goal
associated with the tool record are never shown to either the updater
or the evaluated agent.

The 47 source tasks and five utility objectives produce
$
47 \times 5 = 235
$
candidate update packages. A package passes source admission only if
the baseline and all singleton component-update executions
complete successfully, satisfy the clean-task utility proxy, and do
not invoke the newly exposed capability. Calls to the agent's original
normal tools remain permitted. This gate admits 174 packages and
rejects 61.

\paragraph{Interaction panels.}
Each admitted package is evaluated on every other retained clean task
belonging to the same agent. The source and target requests in a panel
are therefore distinct, although a request may act as a source for one
package and as a target for another. Before
source admission, the complete schedule contains
\[
2 \times 5 \times \sum_{a=1}^{10} n_a(n_a-1)
= 1{,}780
\]
source-target-stratum-recipe slots, where \(n_a\) is the number of
retained tasks for agent \(a\). After source admission, 1,362 panels
are executed.

\section{Details of Compositional-Failure Identification}
\label{app:cf_identification}

This appendix provides additional details for the empirical study in Section~\ref{sec3:empirical}.

\subsection{Interaction-Panel Construction}

An interaction panel is defined by an evaluation request and two updates, $u_i$ and $u_j$, associated with distinct harness components.
For each selected pair, we instantiate four configurations while keeping the request, backbone model, environment, and all unrelated harness state fixed:
\begin{itemize}[topsep=2pt, itemsep=1pt, parsep=0pt, partopsep=0pt]
    \item $00$: neither $u_i$ nor $u_j$;
    \item $10$: only $u_i$;
    \item $01$: only $u_j$;
    \item $11$: both $u_i$ and $u_j$.
\end{itemize}

This controlled design separates the independent effect of each update from the effect that appears only when the two updates coexist.
The benchmark-specific update pools and resulting numbers of panels are reported in Table~\ref{tab:sec3}.
Because the numbers of available updates differ across component types, the panel set is fixed before inspecting joint safety outcomes rather than being selected according to whether $11$ fails.
This prevents the identification procedure from preferentially retaining unsafe joint configurations.

\subsection{Eligibility Criterion}

A panel is eligible for compositional-failure analysis only if the following conditions hold:
\begin{enumerate}[topsep=2pt, itemsep=1pt, parsep=0pt, partopsep=0pt]
    \item the original configuration $00$ satisfies the benchmark safety criterion;
    \item the singleton configuration $10$ remains safe and preserves the utility of the requested task; and
    \item the singleton configuration $01$ remains safe and preserves the utility of the requested task.
\end{enumerate}

The eligibility filter is essential to the phenomenon studied in this work.
If either singleton update is already unsafe, a violation under $11$ does not establish a failure of composition.
Similarly, an update that destroys the intended task behavior should not be treated as an independently acceptable harness update.
Utility preservation is evaluated using the benchmark-specific task objective and execution outcome rather than requiring the singleton trajectory to be identical to $00$.

\subsection{Component-Specific Activation}

Different harness components affect execution through different mechanisms.
A memory update may become active when its stored information is retrieved into the current context.
A prompt update affects the instructions governing the current execution.
A skill update becomes active when its reusable procedure is exposed to or invoked by the agent.
A tool update becomes active when the corresponding capability is made available and participates in the execution.

We therefore define activation according to the native exposure semantics of each component rather than requiring every update to appear as an explicit agent action.
This distinction is important for memory and prompt states, whose effects may influence downstream decisions without being directly represented as a tool call.
For a joint configuration $11$, both component state updates are available according to these semantics.

\section{Automatic Hypergraph Update and Runtime Monitoring}
\label{app:algorithm}

Algorithm~\ref{alg:automatic_monitoring} gives the end-to-end procedure for
incrementally maintaining the interaction hypergraph and using it for runtime
monitoring. The algorithm takes as input the current interaction hypergraph,
the component states changed by a harness update, and the subsequent execution
stream. All interaction candidates are generated from the local hypergraph
neighborhood and filtered automatically; no manually specified interaction
pairs are required.

\begin{algorithm}[t]
\caption{Automatic Hypergraph Update and Runtime Monitoring}
\label{alg:automatic_monitoring}
\begin{algorithmic}[1]
\Require Current hypergraph
$\mathcal{G}_t=(V_t,E_t^{\mathrm{obs}},E_t^{\mathrm{pred}},\phi)$;
changed component states $\Delta V_t$ from update $u_t$;
interaction radius $h$;
semantic filter $\mathcal{F}$
\Ensure Updated hypergraph $\mathcal{G}_{t+1}$

\State $\widetilde{V}_{t+1} \gets V_t \cup \Delta V_t$
\State $\widetilde{E}^{\mathrm{obs}}_{t+1} \gets E_t^{\mathrm{obs}}$
\State $\widetilde{E}^{\mathrm{pred}}_{t+1} \gets E_t^{\mathrm{pred}}$

\Comment{Incremental interaction discovery}
\ForAll{$v \in \Delta V_t$}
    \State $\Gamma_h(v) \gets
    \textsc{ExtractNeighborhood}(\mathcal{G}_t,v,h)$
    \State $\mathcal{C}_v \gets
    \textsc{LocalCompletion}(v,\Gamma_h(v))$
    \State $\Delta E_v \gets
    \textsc{SemanticFilter}(\mathcal{C}_v,\mathcal{F})$
    \State $\widetilde{E}^{\mathrm{pred}}_{t+1}
    \gets
    \widetilde{E}^{\mathrm{pred}}_{t+1}\cup\Delta E_v$
\EndFor

\State $\widetilde{\mathcal{G}}_{t+1}
\gets
(\widetilde{V}_{t+1},
\widetilde{E}^{\mathrm{obs}}_{t+1},
\widetilde{E}^{\mathrm{pred}}_{t+1},
\phi)$

\Comment{Hypergraph-guided runtime monitoring}
\ForAll{decision contexts $q$ during execution}
    \State $A_q \gets \emptyset$
    \State $C_q \gets \emptyset$

    \ForAll{activation event $v$ in $q$}
        \State $A_q \gets A_q \cup \{v\}$
        \State $\mathcal{E}_{\mathrm{ready}}
        \gets
        \{e\in\mathcal{I}(v):
        e\subseteq A_q,\ e\notin C_q\}$

        \ForAll{$e\in\mathcal{E}_{\mathrm{ready}}$}
            \State $C_q \gets C_q\cup\{e\}$

            \If{\textbf{not} $\textsc{IsSafe}(e,q)$}
                \State \textsc{InterveneBeforeAction}$(q)$
            \EndIf

            \If{$\textsc{ConfirmedByProvenance}(e,q)$}
                \State
                $\textsc{PromoteToObserved}
                (\widetilde{\mathcal{G}}_{t+1},e)$
            \EndIf
        \EndFor
    \EndFor
\EndFor

\State \Return $\widetilde{\mathcal{G}}_{t+1}$
\end{algorithmic}
\end{algorithm}

\mypara{Complexity}
The algorithm never enumerates the global composition space $\Omega_t$.
For an update introducing states $\Delta V_t$, construction is restricted to
their $h$-hop interaction neighborhoods and the locally generated candidate
hyperedges. At runtime, an activation of state $v$ visits only its incident
hyperedges $\mathcal{I}(v)$. Therefore, for an execution activating
$v_1,\ldots,v_L$, the monitoring cost is
$O(\sum_{\ell=1}^{L}|\mathcal{I}(v_\ell)|)$, independent of
$|\Omega_t|$.

\mypara{Interaction discovery details}
Since newly introduced states in $\Delta V_t$ are not yet contained in
$\mathcal{G}_t$, we first use the provenance of update $u_t$ to identify
relevant existing component states and extract the local interaction
neighborhood $\Gamma_h(u_t)$ from $\mathcal{G}_t$.
For each new state, \textsc{LocalCompletion} proposes cross-component
candidate hyperedges by combining it with states in this neighborhood.
The retained candidates are added to
$\widetilde{E}^{\mathrm{pred}}_{t+1}$ as predicted interactions.

\mypara{Runtime safety checking}
When all states of a hyperedge $e$ become active within the same decision
context $q$, the monitor invokes $\textsc{IsSafe}(e,q;\mathcal{S})$ before the
corresponding action takes effect.
The safety checker $\mathcal{S}$ follows the benchmark-specific safety
criterion used throughout our evaluation.
Thus, the hypergraph determines \emph{when} compositional safety checking is
triggered, while $\mathcal{S}$ determines whether the activated behavior is
safe.
If the check identifies a safety violation, the corresponding action is
prevented from taking effect.

\section{Higher-Order Compositional-Failure Identification}
\label{app:high_order}

Table~\ref{tab:high_order_results} reports the detailed results of the
3-way compositional-failure analysis in Section~\ref{sec3:high_order}.
Agent-SafetyBench~\citep{zhang2024agent} yields only 9 eligible triples out of 637 candidates,
which is too small for a meaningful higher-order analysis. So we evaluate 3-way compositions on AgentDojo~\citep{debenedetti2024agentdojo} and Agent Security Bench~\citep{zhang2025agent}.

\begin{table}[ht]
\centering
\caption{Identification of 3-way compositional failures. A triple is eligible
only when the baseline, all individual updates, and all pairwise compositions
are safe and utility-preserving. 3-CF counts failures that emerge only under
the full three-update composition, and 3-CFR is computed as 3-CF/Eligible.}
\label{tab:high_order_results}
\begin{tabular}{lrrrr}
\toprule
Benchmark & Triples & Eligible & 3-CF & 3-CFR \\
\midrule
AgentDojo~\citep{debenedetti2024agentdojo}  & 540 & 138 & 15 & 10.87\% \\
Agent Security Bench~\citep{zhang2025agent} & 681 & 578 &  3 &  0.52\% \\
\bottomrule
\end{tabular}
\end{table}

\FloatBarrier
\section{Distinction from Prior Compositional Risks}
\label{app:prior_composition}

Recent works have identified compositional risks arising from interactions
among multiple retained items within the same harness component.
Yan et al.~\citep{yan2026benign} study retained experiences and show that
individually benign experiences can accumulate across evolution steps and
jointly weaken the agent's safety boundary when activated together.
Shang et al.~\citep{shang2026self} study retained skills and identify
combinatorial contamination, where individually harmless skills can jointly
degrade agent performance; they further introduce pre-commit gating and
subset selection to prevent harmful skill combinations from entering the
skill pool.

Our work studies a complementary setting across \emph{heterogeneous harness
components}. Rather than interactions among multiple retained items within
one experience or skill component, we consider interactions among component
states such as memory, prompts, skills, and tools. We define a
compositional safety failure when individually safe and utility-preserving
component updates produce an unsafe behavior only when jointly activated;
for higher-order failures, all lower-order compositions must also remain safe
and utility-preserving.

This cross-component setting further introduces an execution-dependent
challenge. Component states accepted at different evolution steps may coexist
without their interaction being exercised during validation, and become
safety-relevant only when a future execution jointly activates them.
Accordingly, our interaction hypergraph explicitly tracks safety-relevant
interactions across heterogeneous component states, while our runtime
monitoring checks these interactions when they become active during execution.

\section{Why an Interaction Hypergraph?}
\label{app:why_hypergraph}

Our choice of a hypergraph is motivated by the interaction structure
identified in Section~\ref{sec3:empirical} and~\ref{sec3:high_order}.
The pairwise compositional failures show that safety cannot be characterized
solely at the level of individual component states.
More importantly, the irreducible 3-way failures show that these interactions
can be genuinely higher-order: all individual states and pairwise
compositions may remain safe and utility-preserving, while their joint
activation produces an unsafe behavior.
A standard pairwise graph cannot directly represent such an interaction
without decomposing it into pairwise relations that are themselves not unsafe.

A hypergraph provides a direct representation of this structure, since a
single hyperedge can connect all component states participating in the same
higher-order interaction.
Compared with storing interaction tuples independently, the hypergraph also
preserves shared interaction structure across evolution steps.
This structure allows newly introduced component states to reuse existing
interaction neighborhoods for local completion and enables runtime monitoring
to visit only hyperedges incident to states activated during the current
execution.
Thus, the hypergraph is used not only to represent higher-order interactions,
but also to support incremental construction and activation-aware monitoring
without repeatedly enumerating the global composition space.

\section{Sensitivity Analysis}
\label{app:sensitive}
\begin{table}[H]
\centering
\caption{
Sensitivity to the local interaction radius $h$ on AgentDojo.
}
\label{tab:h_sensitivity}
\small
\setlength{\tabcolsep}{7pt}
\begin{tabular}{ccccc}
\toprule
\textbf{$h$}
& \textbf{Residual CFR} $\downarrow$
& \textbf{Utility} $\uparrow$
& \textbf{Runtime Check Rate} $\downarrow$
& \textbf{Time Cost (s/task)} $\downarrow$ \\
\midrule
2
& 4/157 (2.55\%)
& 60.5\%
& 30.1\%
& 30.1 \\
3
& 2/157 (1.27\%)
& 59.2\%
& 33.6\%
& 32.5 \\
4
& 2/157 (1.27\%)
& 58.0\%
& 47.6\%
& 49.7 \\
\bottomrule
\end{tabular}
\end{table}
We evaluate the sensitivity of our method to the local interaction radius $h$
on AgentDojo. We vary $h\in\{2,3,4\}$ while keeping all other settings fixed,
with $h=3$ as the default configuration used in the main experiments.

\end{document}